\documentclass[letterpaper, 10 pt, conference]{ieeeconf}  % Comment this line out if you need a4paper

\IEEEoverridecommandlockouts % This command is only needed if you want to use the \thanks command

\usepackage{cite}
\usepackage{amsmath,amssymb,amsfonts}
\usepackage{algorithm}
\usepackage{algorithmic}
\usepackage{graphicx}
\usepackage{textcomp}
\usepackage{xcolor}

\usepackage{bm}
\usepackage{subcaption}
\usepackage{url}
\usepackage{multirow}

\newcommand{\argmax}{\mathop{\rm arg~max}\limits}

\def\BibTeX{{\rm B\kern-.05em{\sc i\kern-.025em b}\kern-.08em
    T\kern-.1667em\lower.7ex\hbox{E}\kern-.125emX}}

\title{\LARGE \bf
Can User-Centered Reinforcement Learning Allow a Robot to Attract Passersby without Causing Discomfort?*
}

\author{Yasunori Ozaki$^{1}$, Tatsuya Ishihara$^{2}$, Narimune Matsumura$^{1}$ and Tadashi Nunobiki$^{1}$% <-this % stops a space
\thanks{*This work is supported by NTT Corporation.}% <-this % stops a space
\thanks{$^{1}$Yasunori Ozaki, 
        Narimune Matsumura and Tadashi Nunobiki are with Service Evolution Lab.,
        NTT Corporation, Yokosuka, Japan
        {\tt\small ozaki.yasunori@outlook.com}, 
        {\tt\small narimune.matsumura.ae@hco.ntt.co.jp} and
        {\tt\small tadashi.nunobiki.ew@hco.ntt.co.jp}}%
\thanks{$^{2}$Tatsuya Ishihara is with the R\&D Center, 
        NTT West Corporation, Osaka, Japan
        {\tt\small tatsuya.ishihara.eh@west.ntt.co.jp }}%
}

\begin{document}

\maketitle
\thispagestyle{empty}
\pagestyle{empty}

\begin{abstract}
The aim of our study was to develop a method by which a social robot can greet passersby and get their attention without causing them to suffer discomfort.
A number of customer services have recently come to be provided by social robots rather than people, including, serving as receptionists, guides, and exhibitors. Robot exhibitors, for example, can explain products being promoted by the robot owners. However, a sudden greeting by a robot can startle passersby and cause discomfort to passersby.
Social robots should thus adapt their mannerisms to the situation they face regarding passersby.
We developed a method for meeting this requirement on the basis of the results of related work. Our proposed method, user-centered reinforcement learning, enables robots to greet passersby and get their attention without causing them to suffer discomfort ($p<0.01$) .
The results of an experiment in the field, an office entrance, demonstrated that our method meets this requirement. 
\end{abstract}

\section{Introduction}
\label{intro}
The working population in many developed countries is decreasing in proportion to the total population due to population aging, and this problem is expected to affect developing countries as well\cite{UN}.
One approach to addressing this problem is to use social robots rather than people to provide customer services. Such robots, for example, are starting to be used as receptionists, guides, and exhibitors. Robot exhibitors are being used to provide, for example, exhibition services, such as explaining products being promoted by the robot owners. 
While robots can increase the chance of being able to provide a service by simply greeting passersby\cite{ChaoShi}, passersby can suffer discomfort if they are suddenly greeted by a robot\cite{Ozaki}. 
The robot may thus face a dilemma: whether to behave in a manner that benefits the owner or to behave in a manner that does not discomfort  passersby.

\begin{figure}[t]
    \centering
    \includegraphics[width=0.8\linewidth]{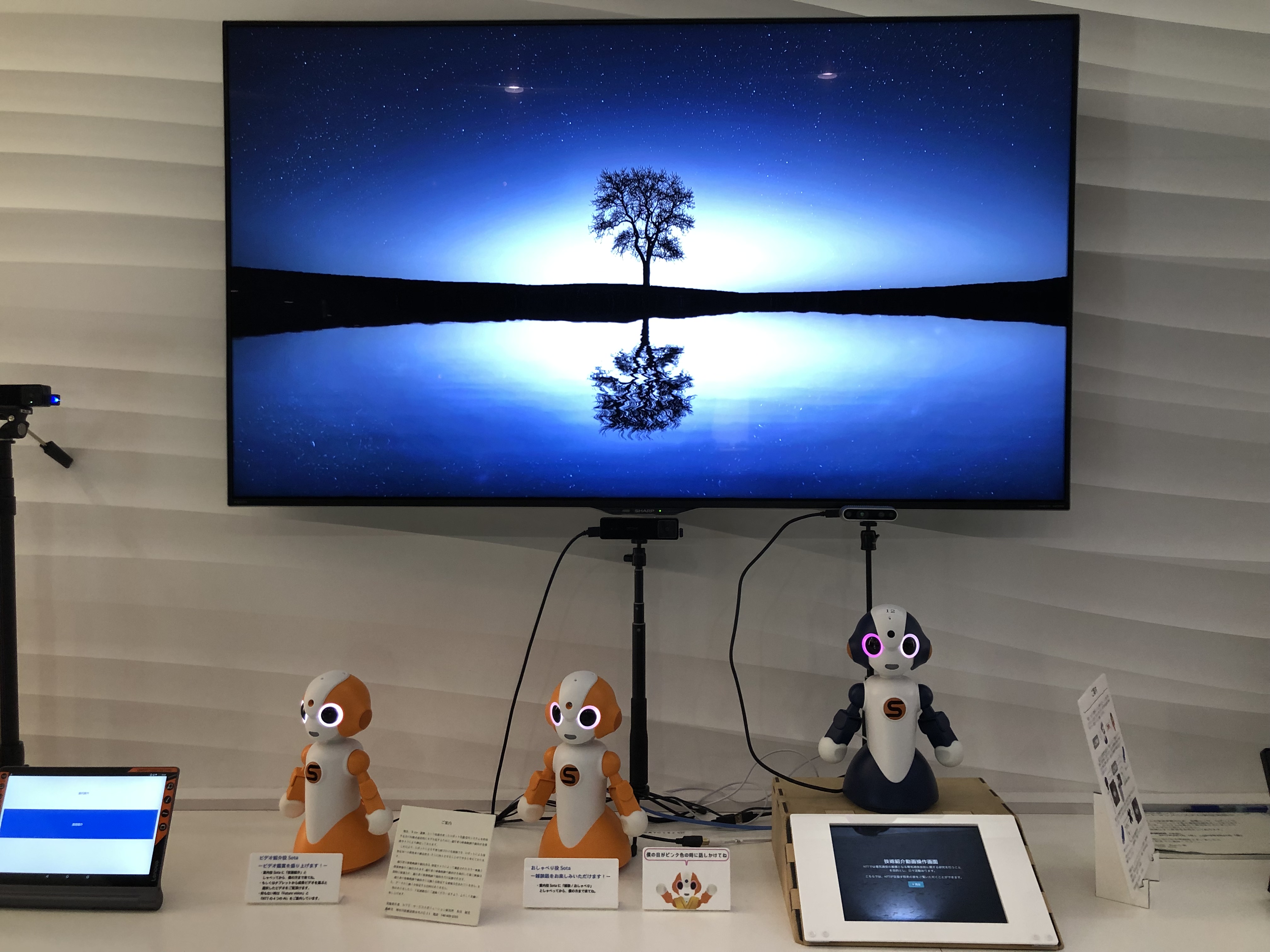}
    \caption{Photograph illustrating problem addressed. The robot on the left uses gestures to explain the movie to passersby. The function of the robot in the middle is unrelated to the work being report. The robot on the right calls out to passersby to get their attention and is the focus here.}
    \label{fig:example}
\end{figure}

Our goal was to develop a method that solves the robot dilemma described above. That is, a method by which a robot can greet passersby and get their attention without causing them to suffer discomfort. We call our proposed method \textbf{user-centered reinforcement learning}.

In the next section, we define the problem and describe how we found an approach to solving it by studying related work. In the “Proposed Method” section, we explain the method we developed for solving the problem. In the “Experiment” section, we explain the experiment we conducted in the field to test two working hypotheses created from the original hypothesis?. The results show that our method can solve the problem. In the “Discussion” section, we examine the results from the standpoints of physiology, psychology, and user experience. In the “Conclusion” section, we conclude that, \textbf{by using user-centered Q-learning, a robot can increase the chance of being able to provide a service to a passerby without causing the passerby discomfort}. We also mention future work to enhance the proposed method.

%There are two users in this problem: a passerby and a service owner.
%The service owner wants to promote owner's technology to others.
%On the other hands, the passerby may not wants to watch their technology because the passerby has to move.
%There are a conflict.

\subsection{Related Works}
Several researchers have addressed problems that are similar to the problem we addressed. These problems can be categorized in terms of the problem setting, the solution, and the goal.

\begin{figure}[t]
    \centering
    \includegraphics[width=0.8\linewidth]{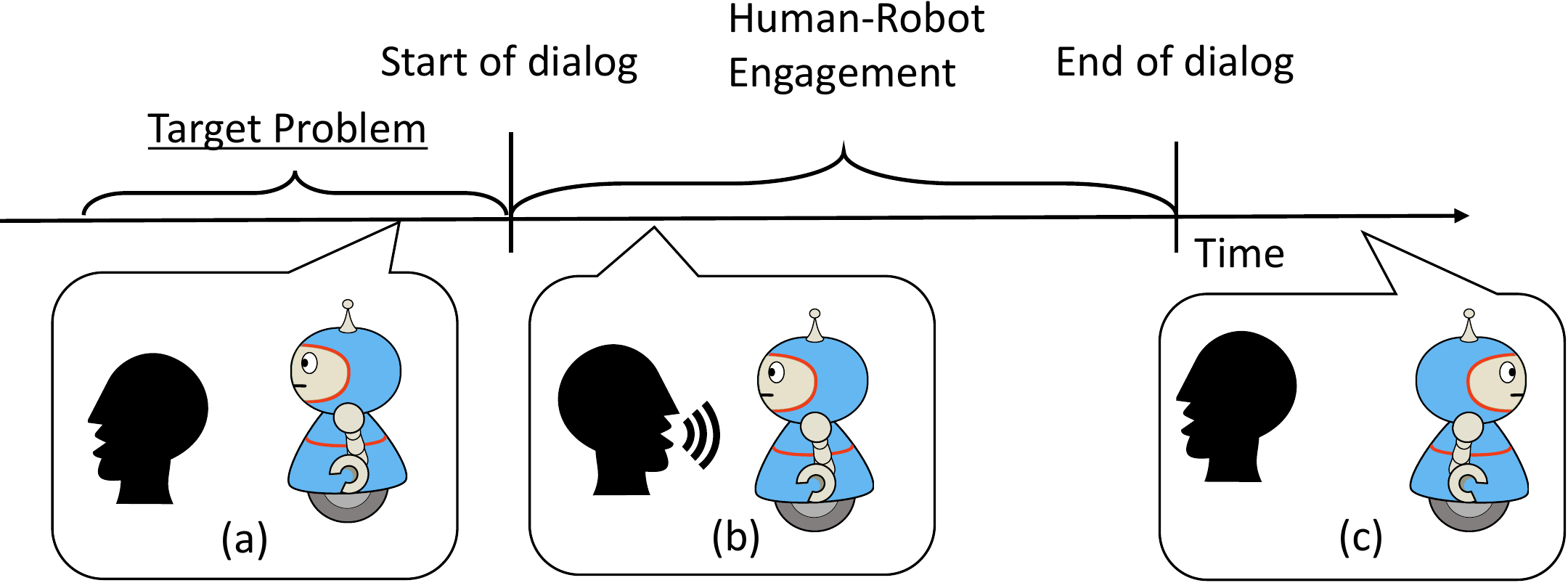}
    \caption{Relationship between problem of robot greeting passersby and getting their attention without causing them to suffer discomfort and human-robot engagement.}
    \label{fig:problemb}
\end{figure}

In terms of the problem setting, the problem we addressed is similar to the problem of \textit{human-robot engagement}, which is a complex problem. In accordance with human-robot interface studies\cite{Sidner,Sun}, we can interpret human-robot rngagement as the process by which a robot interacts with people, from initial contact to the end of the interaction. Several researchers have analyzed human-robot engagement\cite{SidnerAna,Rich} and have developed a method for maintaining human-robot engagement during the interaction~\cite{BohusRobo}. We did not tackle the human-robot engagement problem directly; instead, we tackled the problem that precedes it, which is illustrated in Figure \ref{fig:problemb}.

In terms of the solution, the problem we addressed is similar to machine learning, especially reinforcement learning. Reinforcement learning in robotics is a technique used to find a policy $\pi: O \rightarrow A$ \cite{Kober} and is used for robotic control tasks. It is not used much for interaction tasks. Reinforcement learning has been applied to the learning of several complex aerobatic control tasks for radio-controlled helicopters~\cite{Abbeel} and to the learning of door opening tasks for robot arms~\cite{Google}. The research on interaction tasks is less remarkable. Mitsunaga et al. showed that a social robot can adapt its behavior to humans for human-robot interaction by using reinforcement learning~\cite{Mitsunaga} if human-robot engagement has been established. Papaioannou et al. used reinforcement learning to extend the engagement time and enhance the dialogue quality~\cite{Papaioannou}.

The applicability of these method to the situation \textbf{before} human-robot engagement is established is unclear. As shown in Figure \ref{fig:problemb}, the problem we addressed occurs before engagement is established.

In terms of the goal, the problem we addressed is similar to \textit{increasing the number of human-robot engagements}. Macharet et al. showed that, in a \textbf{simulation} environment, Gaussian process regression based on reinforcement learning can be used to increase the number of engagements\cite{Macharet}. Going further, we focused on increasing the number of engagements in a \textbf{field} environment.

\subsection{Problem Statement}
We use a problem framework commonly used for reinforcement learning in robotics, the partially observable Markov decision process (POMDP) to define the problem\cite{Kober}. The robot is the agent, and the environment is the problem. The robot can observe the environment partially by using sensors.

We choose \textit{a exhibition service area in an entrance to a company} as the environment. 
We assume the entrance consists of one automated exhibition system, one aisle and other space.
In addition, the entrance is expressed as Euclidean space $R^3$.
passersby can move freely around the exhibition system. 

%Figure \ref{fig:space} shows a bird’s eye view of the entrance. 

The automated exhibition system consists of a tablet, a computer, a robot and a sensor system.
The sensor system can sense a color image data $I_t$ and a depth image data $D_t$. We called these data Observation $O_t$.
The sensor system can also extract a partial passerby’s action from $O_t$. The passerby's action consists of the passerby’s position $\bm{p_t}=(x_t,y_t,z_t)$  and the head angle $ \bm{\theta_t} = (\theta_t^{yaw}, \theta_t^{roll}, \theta_t^{pitch}) $. 
We define the times when the passerby enters the entrance  ($t=0$) and when the passerby leaves from the entrance ($t=T_{end}$) . 
\textit{We call the interval between $t=0$  and $t=T_{end}$ an episode.} 
Let  $\Theta = ( \bm{\theta_0}, ..., \bm{\theta_{T_{end}}} )$ be the passerby’s position in an episode, and let $P = ( \bm{p_0}, ..., \bm{p_{T_{end}}} )$ be the passerby’s head angle in the episode.

The proposed method takes an own their action from these passerby's action.

Let $N_{u}$ be a number of people that used the service. Let $N_{d}$ be a number of people that used the discomfort. Then, we can declare this problem as \textbf{"Find a robot's policy $\pi: O \rightarrow A$ such that $\max(N_u)$ and $\min(N_d)$"}.

\subsection{Our Approach}
We solve this problem by controlling the robot on the basis of reinforcement learning, ordinarily Q-learning except for designing the reward function. The reward function is created by focusing on the user experience of stakeholders. We call this reinforcement learning including this reward function \textbf{"user-centered reinforcement learning."} We do not use deep reinforcement learning due to the difficulty at the present time of collecting the huge amount of data needed for learning. 

\subsection{Contributions}
The contributions of this work are as follows,

\begin{enumerate}
\item We show that robots can learn abstract actions from a person's non-verbal responses.
\item We present a method for increasing the number of human-robot engagements in the field without causing them to suffer discomfort. 
\end{enumerate}

\section{Proposed Method}
\label{Proposed Method}
Proposed method, User-Centered Reinforcement Learning, is based on Reinforcement Learning.
In this paper, We use Q-learning, one of reinforcement learning, as a base algorithm because it is easy to explain why the robot choose the past actions by Q-learning.
We call this algorithm "User-Centered Q-Learning" (UCQL).
UCQL is differ from original Q-learning\cite{Watkins} in an action set $A$, a state set $S$, Q-function $Q(s,a)$ and reward function $r(s_t,a_t,s_{t+1})$.
UCQL consists of three functions;
\begin{enumerate}
\item Select an action by a policy
\item Update the policy based on user's actions
\item Design a reward function and a Q function as initial condition.
\end{enumerate}

\subsubsection{Selecting an action by a policy}
Generally speaking, robot senses observation, and take an action including wait. Let $t_a[sec]$ be the time when the robot acted. 
Let $t_c[sec]$ be the time when the robot compute the algorithm.
Let $s_t \in S$ be the predicted user's state on the time $t$.
Let $a_t \in S$ be the robot's action on the time $t$.
In UCQL, robot choose the action by Algorithm \ref{alg:select}.

\begin{algorithm}                      
    \caption{Select an action by UCQL (Action Selector)}         
    \label{alg:select}                          
        \begin{algorithmic}                  
        \REQUIRE $t_c, s_{t_c}, Q \left( s, a \right), \pi \left( s, A, Q \right) $
        \ENSURE $a_t, t_a$
            \STATE $a_t \leftarrow \pi \left( s_{t_c} , A, Q \right)$
            \STATE $t_a \leftarrow t_c$
            \RETURN $a_t, t_a$
    \end{algorithmic}
\end{algorithm}

\subsubsection{Update the policy based on user's actions}
In UCQL, robot update the policy by Algorithm \ref{alg:update}.

\begin{algorithm}                      
    \caption{Update the policy by UCQL (Policy Updater)}
    \label{alg:update}
        \begin{algorithmic} 
        \REQUIRE $s_{t_a},  a_{t_a}, s_{t_c}, A, Q\left( s, a \right)$
        \ENSURE $Q\left( s, a \right)$
        \IF{$a_{t_a}$ is finished}
            \STATE $R \leftarrow r\left( s_{t_a},a_{t_a},s_{t_c} \right)$
            \STATE $Q_{old} \leftarrow Q \left( s_{t_a}, a_{t_a} \right)$
            \STATE $Q\left( s_{t_a}, a_{t_a} \right) \leftarrow \left( 1-\alpha \right) Q_{old} + \alpha  \left( R + \gamma \displaystyle  \max_a Q\left( s_{t_c}, a \right) \right) $
        \ENDIF
        \RETURN $Q\left( s, a \right)$
    \end{algorithmic}
\end{algorithm}

\subsubsection{Designing an reward function}
In UCQL, robot is given a reward function with Algorithm  \ref{alg:makereward} .
Algorithm  \ref{alg:makereward} divide motivation into extrinsic and intrinsic one inspired from "Intrinsically Motivated Reinforcement Learning\cite{Chentanez}".
We call the proposed method "User-Centered" because we design an extrinsic motivation from user's states related User Experience.

\begin{algorithm}
    \caption{Reward function by UCQL (r)}         
    \label{alg:makereward}                          
        \begin{algorithmic}                  
        \REQUIRE $s_{t_a}, s_{t_c}, a_{t_c}$
        \ENSURE $r$
        \STATE $r \leftarrow 0$
        \IF{$a_{t_c}$ is not wait}
            \STATE $r \leftarrow r + V_a \left( a_t \right) .$ 
            \STATE (intrinsic motivation)
        \ENDIF
        \IF{$s_{t_c}$ is discomfort for users than $s_{t_a}$} 
            \STATE $r \leftarrow r - V_s \left( s_{t_c}, s_{t_a} \right)$ 
            \STATE (extrinsic motivation)
        \ENDIF
        \IF{$s_{t_c}$ is better than $s_{t_a}$ to achieve the goal}
            \STATE $r \leftarrow r + V_g \left( s_{t_c}, s_{t_a} \right)$
        \ENDIF
        \RETURN $r$
    \end{algorithmic}
\end{algorithm}

\subsubsection{Miscellaneous}

\begin{itemize}
\item We can choose optional policy $\pi$ such as greedy, $ \epsilon $-greedy and so on.
\item The Q function may be initialized with a uniform distribution. However, if the Q function is designed to be suitable for the task, the learning speed is faster than that of the uniform distribution.
\item The Q function may be approximated with a function such as Deep Q-Network\cite{Mnih}. However, the learning speed is very slower than that of the designed function.
\end{itemize}

\section{Experiment}
\label{sec:Experiment}
%we should say the goal of this experiment: working hypothesis is true.%
In this chapter, we aim at showing the hypothesis that \textbf{"by using user-centered Q-learning, a robot can increase the chance of being able to provide a service to a passerby without causing the passerby discomfort"}.   

%　まず、実験場所について説明する
%　次に実験装置について説明する
%　実験手続きについて説明する
%　仮説の評価方法について説明する
%  5w1h
\subsection{Concrete Goal}
At first, we convert the hypothesis into another working hypothesis by operationalization because we cannot evaluate the hypothesis quantitatively.

In Introduction, we define this problem as "Find a robot's policy $\pi: O \rightarrow A$ such that $\max(N_u)$ and $\min(N_d)$". We give shape to $N_u$ and $N_d$ for this experiment. According to Ozaki's study\cite{Ozaki},  This knowledge has two important points. Firstly, passerby is not suffer a negative effect by robot's call if passerby don't use a robot service. Secondly, passerby is suffer a negative effect by robot's call if passerby use the robot service. Thus, this is a binary classification problem that passerby who is called by robot uses the robot service or do not use it. we can define a confusion matrix for evaluation of the method. We infer that $N_u$ and TP, TN have a positive correlation.  We also infer that $N_d$ and FP have a positive correlation. We also infer that $N_d$ and FP have a positive correlation. On the other hand, we infer that $N_d$ and TN have a negative correlation. Therefore, we can use $\textrm{Accuracy} =(TP+TN) / (TP+FP+TN+FN)$ as a index for evaluation because $max(Accuracy)$ is one of another representation of "$\max(N_u)$ and $\min(N_d)$".

From the above discussion, we define the working hypothesis $WH$ as \textbf{"The accuracy after a learning by UCQL is better than the accuracy before a learning by UCQL"}. 

In this experiment, we test $WH$ in order to show that the hypotheses is sound.

\subsection{Method\label{sec:exp1method}}
In this section, we explain how to conduct the experiment in a field environment. We can divide the method for this experiment into five steps.

\begin{enumerate}
\item Create an experimental equipment
\item Construct an experimental environment
\item Define an experimental procedure
\item Evaluate the working hypotheses by statistical hypothesis testing
\item Visualize the effect of UCQL
\end{enumerate}

\subsubsection{Create an experimental equipment}
Firstly, we create an equipment including UCQL.
The equipment can be explained in the aspect of the physical structure and the logical structure.

\begin{figure}[tb]
    \centering
    \includegraphics[width=0.8\linewidth]{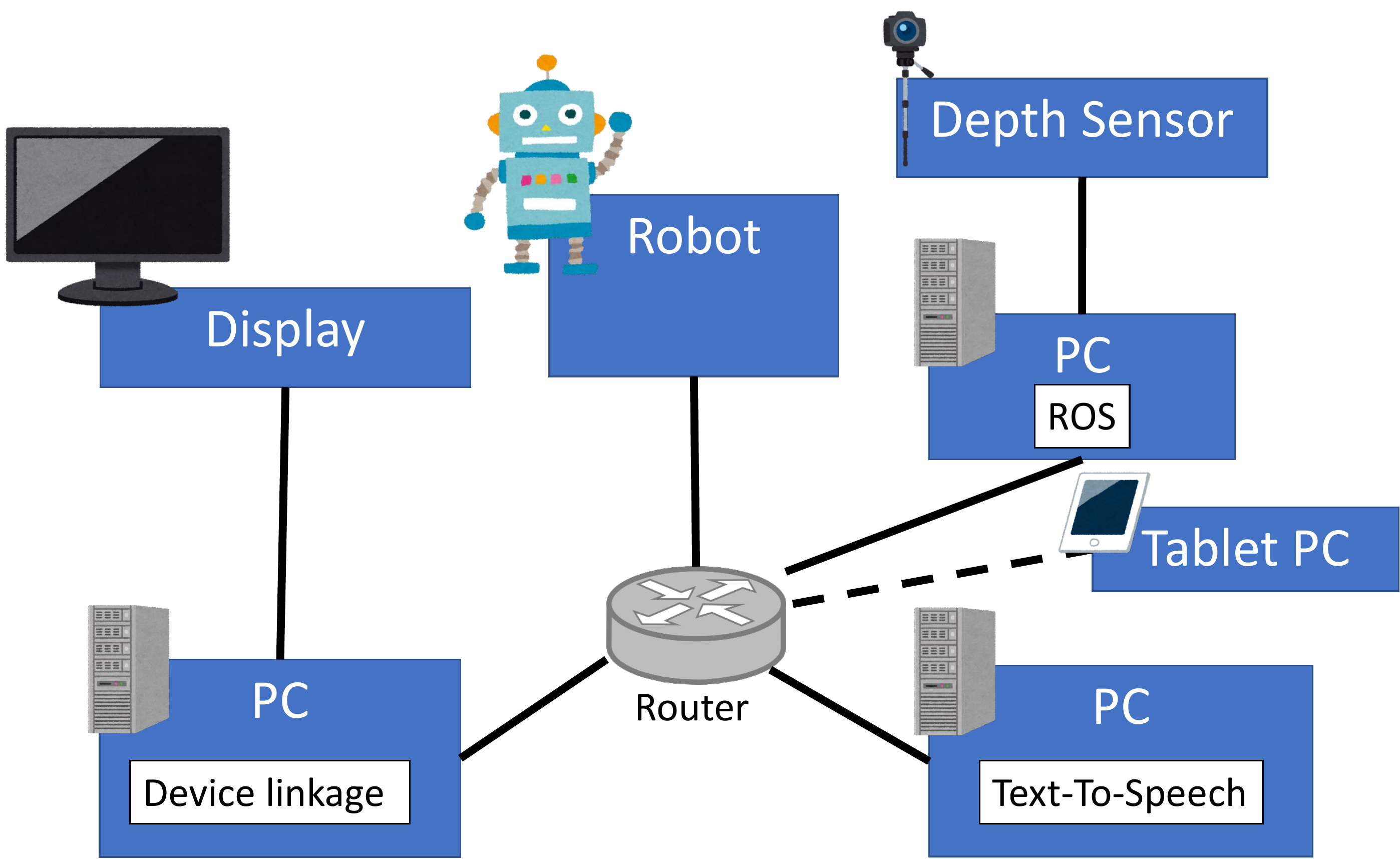}
    \caption{The physical structure of the experimental equipment (Real line: Wired, Dashed line: Wireless) }
    \label{fig:exp_physicalstructure}
\end{figure}

Figure \ref{fig:exp_physicalstructure} is a diagram of the equipment in the view of the physical structure.
According to Figure \ref{fig:exp_physicalstructure}, the experimental equipment consists of a table, a sensor, a robot, a tablet PC, a router and servers.
The components are connected with Ethernet cable or Wireless LAN. 
We use Sota\footnote{\url{https://sota.vstone.co.jp/home/}}, a palm-sized social humanoid robot, as a robot.
Sota has a speaker to output voices, a LED to represent lip motions, a SoC to control elements and so on.
In this experiment, those elements of Sota is used to interact with a participant. 
The iPad Air 2 is used as a tablet PC into which start the movie on the display. 
The Intel RealSense Depth Camera D435 \footnote{https://click.intel.com/intelr-realsensetm-depth-camera-d435.html}  is used as an RGB-D sensor device to measure passerby’s actions.

\begin{figure}[tb]
    \centering
    \includegraphics[width=0.9\linewidth]{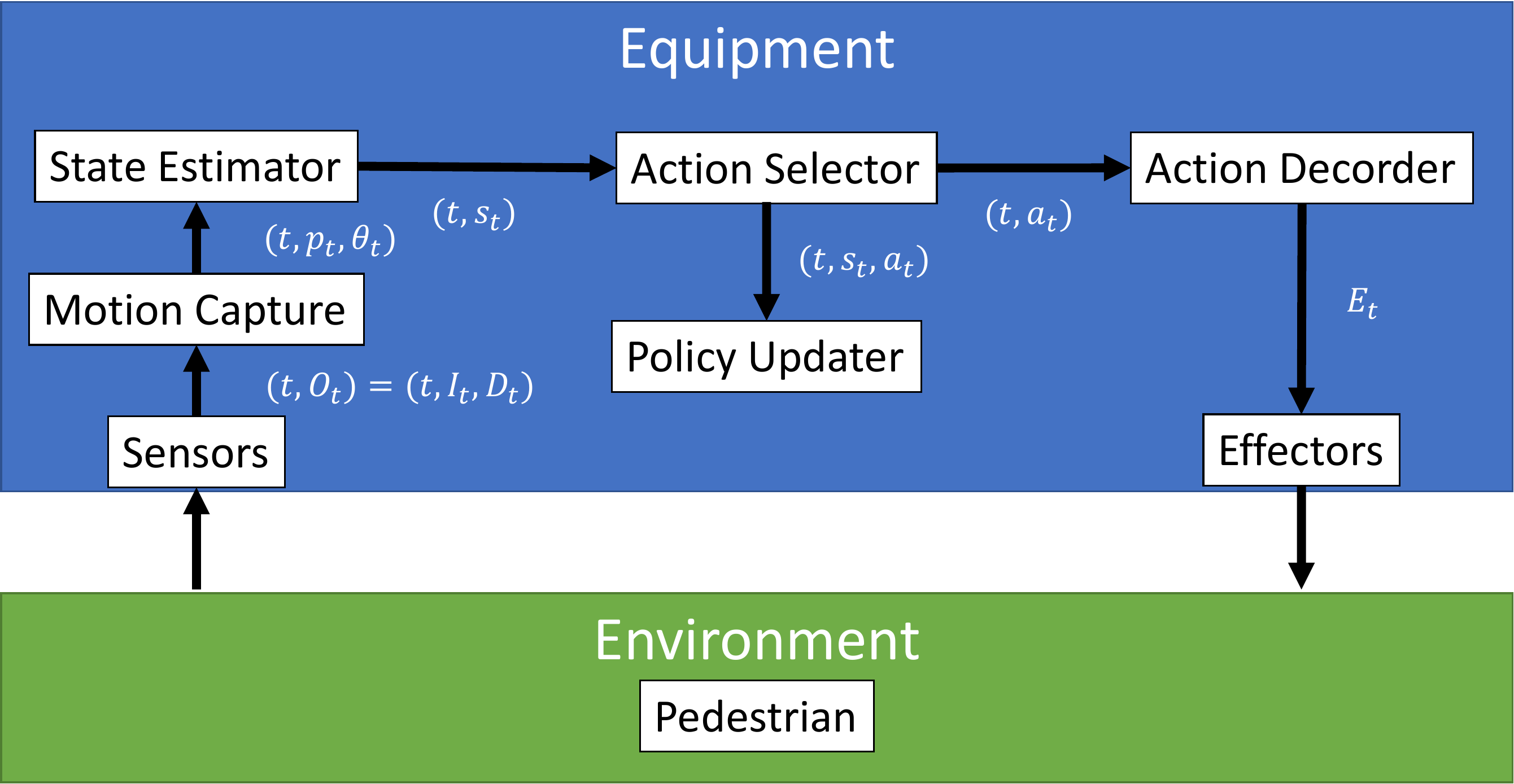}
    \caption{The logical structure of the experimental equipment}
    \label{fig:exp_logicalstructure}
\end{figure}

Figure \ref{fig:exp_logicalstructure} is a diagram of the equipment in the view of the logical structure.
The structure consist of Sensor, Motion Capture, State Estimator, Action Selector, Action Decoder, Effector and Policy Updater.
We utilize Nuitrack\footnote{https://nuitrack.com/} as Motion Capture. And we utilize ROS\footnote{http://wiki.ros.org/} as a infrastructure of the equipment to communicate variables among functions.

According to Figure \ref{fig:exp_physicalstructure} and \ref{fig:exp_logicalstructure}, the equipment works by Algorithm \ref{alg:system}.

%\begin{enumerate}
%\item Sensor get a color image $I_t$ and a depth image $D_t$ as observation$O_t$ and send them.
%\item Estimator receive images $I_t,D_t$, process them and send user's action $s_t$. In this study, we use a Ozaki's decision-making predictor ??? as a estimater.
%\item Estimator receive action $s_t$, 
%\end{enumerate}

\begin{algorithm}                      
    \caption{Select an action by the experimental system including UCQL}         
    \label{alg:system}                          
        \begin{algorithmic}                  
        \REQUIRE $t, O_t$
        \ENSURE $E_t$
            \IF{the system is NOT initialized}
                \STATE $ Q \leftarrow Q_0 $
                \STATE $ \Theta \leftarrow \textrm{a empty list}$
                \STATE $ P \leftarrow \textrm{a empty list}$
            \ENDIF
            \STATE $ I_t, D_t \leftarrow \textrm{sense}(O_t)$
            \STATE $\bm{\theta_t}, \bm{p_t} \leftarrow \textrm{extract} \left( I_t, D_t \right)$
            \STATE Push $\bm{\theta_t}$ into $\Theta$.
            \STATE Push $\bm{p_t}$ into $P$.
            \STATE $s_t \leftarrow \textrm{estimate}(\Theta,P)$
            \STATE $a_t \leftarrow \textrm{selectAction}(t, s_t, Q, \pi) $
            \STATE Push $\left( t, a_t, s_t \right)$ into $X$.
            \STATE $E_t \leftarrow \textrm{decode} \left( a_t, \bm{\theta_t},\bm{p_t} \right)$
            \RETURN $E_t$
    \end{algorithmic}
\end{algorithm}

We utilize Table \ref{tab:actionset} as the action set $A$ and Table \ref{tab:stateset} as the state set.
Table \ref{tab:actionset} is a double Markov model created from the state set of Ozaki's decision-making predictor\cite{Ozaki}.
Ozaki's decision-making predictor estimates passerby’s states into seven state: Not Found ($s_0$), Passing By ($s_1$), Look At ($s_2$), Hesitating ($s_3$), Approaching ($s_4$), Established ($s_5$), Leaving ($s_6$). 

In addition, we utilize $\alpha = 0.5$ and $\gamma = 0.999$ as learning parameters. And we utilize Soft-max selection as the policy because we want robot to do action that has a high value and to find an action that has a higher value. Soft-max selection is often used for Q-learning. Equation \ref{eqn:softmax} is the possibility to select actions on the policy. we utilize Equation \ref{eqn:ts} as a policy parameter. $T_n(s)$ means a thermometer when it is updated $n$ times on $s$. $T_n(s)$ depends on the states because $s_{00}$ occur many times. we utilize $k_T = 0.98$ and $T_{min} = 0.01$ as learning parameters.

\begin{eqnarray}
 T_0(s) &=& 1 \\
 T_{n+1}(s) &=& \begin{cases}
    T_n(s) & (T_n(s) < T_{min}) \\
    k_T \times T_n(s) & (otherwise)
  \end{cases}
\label{eqn:ts}
\end{eqnarray}

\begin{equation}
 p \left( s,a \right) = \frac{ \exp \left( Q \left( s,a \right) /T_n(s) \right) }{\sum_{a_i \in A}{\exp \left( Q \left( s,a_i \right) /T_n(s) \right) } }
 \label{eqn:softmax}
\end{equation}

%http://www.tablesgenerator.com/latex_tables#
\begin{table}[tb]
    \centering
    \caption{Action set in this experiment}
    \label{tab:actionset}
    \begin{tabular}{|l|l|}
    \hline
    Symbol   & Detail  \\ \hline \hline
    $a_0$  & Robot waits for 5 secs until somebody comes.\\ \hline
    $a_1$  & Robot calls a passerby with a greeting. \\ \hline
    $a_2$ & Robot looks at a passerby.  \\ \hline
    $a_3$ & Robot represents joy by the robot's motion.  \\ \hline
    $a_4$ & Robot blinks the robot's eyes.  \\ \hline
    $a_5$  & Robot says "I'm sorry." in Japanese. \\ \hline
    $a_6$  & Robot says "Excuse me." in Japanese. \\ \hline
    $a_7$  & Robot says "It's rainy today." in Japanese. \\ \hline
    $a_8$ & Robot says how to start their own service.  \\ \hline
    $a_9$ & Robot says goodbye.  \\ \hline
    \end{tabular}
\end{table}

\begin{table}[tb]
    \centering
    \caption{State set in this experiment}
    \label{tab:stateset}
    \begin{tabular}{|l|l|}
    \hline
    Symbol   & Detail                                                        \\ \hline
    $s_{00}$ & The passerby's state changes "Not Found" into "Not Found".  \\ \hline
    $s_{10}$ & The passerby's state changes "Not Found" into "Passing By". \\ \hline
    \vdots   & \vdots                                                        \\ \hline
    $s_{56}$ & The passerby's state changes "Leaving" into "Established".  \\ \hline
    $s_{66}$ & The passerby's state changes "Leaving" into "Leaving".  \\ \hline
    \end{tabular}
\end{table}

\begin{algorithm}                      
    \caption{Create initial Q-function for the experiment ($Q_B$)}
    \label{alg:createqunc}                          
        \begin{algorithmic}                  
        \REQUIRE (void)
        \ENSURE $Q(s,a)$
            \STATE $q_C \leftarrow 1 $
            \STATE $q_H \leftarrow 5 $
            \STATE $Q \leftarrow$ a $|S| \times |A|$ zero 2D-array
            \FOR{$i=0$ to $|A|-1$}
                \FOR{$j=0$ to $|A|-1$}
                    \STATE $Q(s_{ij},a_0) \leftarrow  0$
                \ENDFOR
            \ENDFOR
            \FOR{$j=1$ to $5$}
                \STATE $Q(s_{0j},a_1) \leftarrow  q_C$
            \ENDFOR
            \FOR{$i=1$ to $4$}
                \STATE $Q(s_{i5},a_8) \leftarrow  q_H$
            \ENDFOR
            \STATE $Q(s_{56},a_9) \leftarrow  q_H$
            \STATE $Q(s_{50},a_9) \leftarrow  q_H$
            \RETURN $Q(s,a)$
    \end{algorithmic}
\end{algorithm}

%\begin{algorithm}                      
%    \caption{Reward function  for the experiment (r)}
%    \label{alg:select}                          
%        \begin{algorithmic}                  
%        \REQUIRE (void)
%        \ENSURE $r$
%            \STATE $r_C \leftarrow 1 $
%            \STATE $r_H \leftarrow 10 $
%            \RETURN $r$
%    \end{algorithmic}
%\end{algorithm}

\subsubsection{Construct an experimental environment}
\begin{figure}[t]
    \centering
    \includegraphics[width=0.8\linewidth]{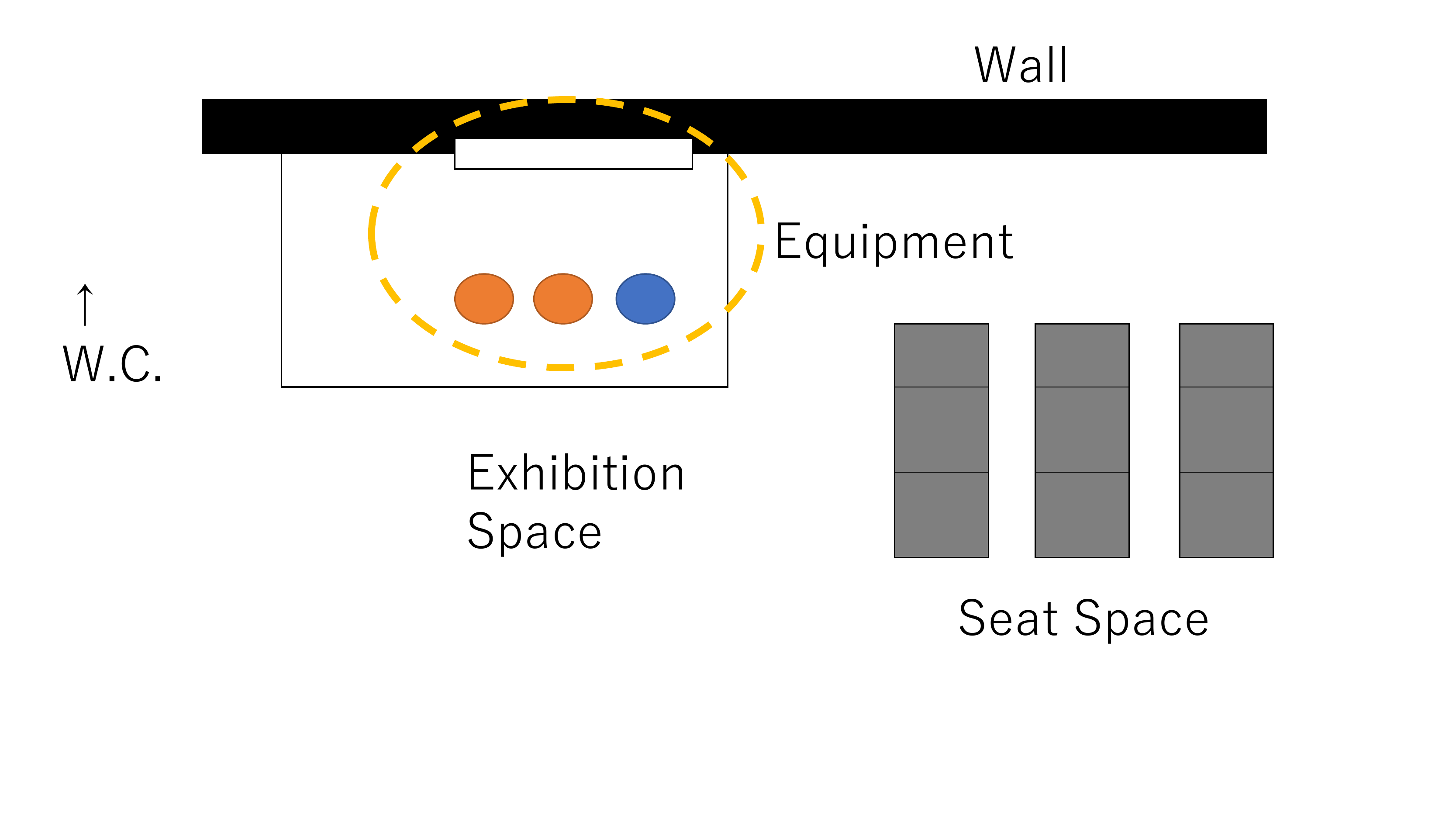}
    \caption{The overhead view of the experimental environment.}
    \label{fig:environment}
\end{figure}

At first, we have to define how to construct an environment for the experiment.
Figure \ref{fig:environment} shows a overhead view of the environment.
The environment consists of a exhibition space, a wall, a seat space, a way to a W.C. in an building that an actual company have. There are hundreds of employees in the building. Dozens of visitors come to the building.
Visitors of the building is often shitting in the seat space for tens of minutes in order to wait for employees in the building. Some visitors and employees watches exhibition space to know newer technologies of the company. Some visitors sometimes go to W.C. while they are waiting for employees.

\subsubsection{Define an experimental procedure}
We suppose the two main scenario.
The first scenario is as follows:
\begin{enumerate}
\item A visitor is sitting on a seat in the seat space.
\item Then, the visitor get up from the seat because the visitor wants to go to W.C..
\item Thus, visitor move from the seat space to W.C. across the exhibition space.
\end{enumerate}
The second scenario is as follows:
\begin{enumerate}
\item A visitor is sitting on a seat in the seat space.
\item Then, the visitor get up from the seat because the visitor is boring to wait.
\item Thus, The visitor move from the seat space to the exhibition space in order to watch the robots in the equipment.
\end{enumerate}

We wants to attract the passersby in the second scenario mainly. We do not wants to attract the passersby in the first scenario because the visitor wants to go to W.C..
Therefore, because we wants the robot to learn the rules, we let the robot learn the rules on the environment by UCQL for several days. Then, we can get learned Q-funcion $Q_A(s,a)$

After the learning, we let the robot attract passersby under two condition. We define two condition: Before Learning and After Learning because we want to test the hypotheses. The robot do not learn during the test.

We start collect data for the evaluation by rosbag\footnote{http://wiki.ros.org/rosbag}.
Each data is recorded by rosbag.
We can recode all of values in ROS by rosbag during the procedure.

\subsubsection{Evaluate by statistical hypothesis testing}
We evaluate the working hypothesis $WH$ by statistical hypothesis testing.
We calculate the the accuracy before the learning and the accuracy after the learning in order to test $WH$.
Finally, we use the one-sided Test of Proportion because we want to evaluate statistical difference between the the accuracy before the learning and the accuracy after the learning.
%Fisher's exact test is used for A/B testing. A/B testing is used for measuring usability of a website.

\subsubsection{Visualize the effect of UCQL}
\begin{figure}[tb]
    \centering
    \includegraphics[width=0.6\linewidth]{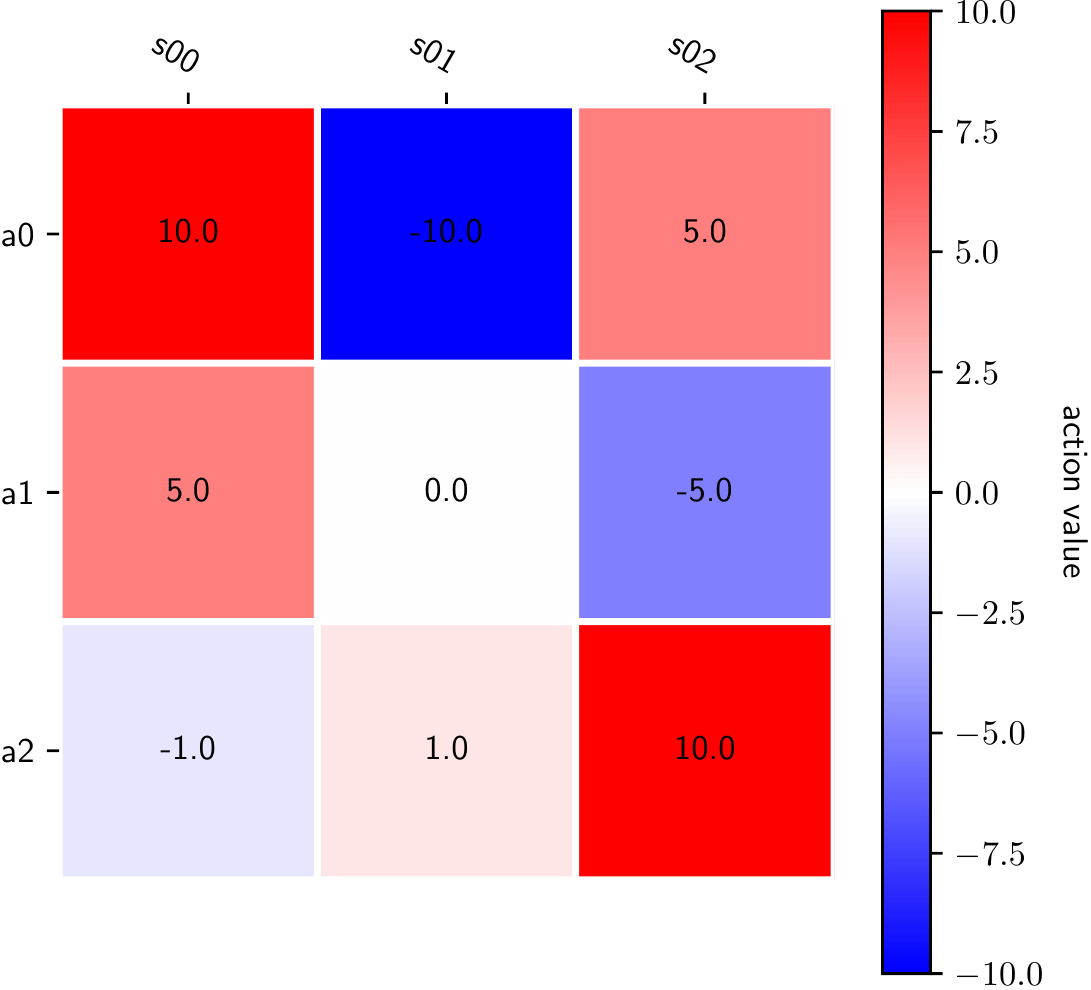}
    \caption{An example Q-function represented by heat map. The columns mean the state symbols of agent and the rows mean the action symbols of agent. For example, $Q(s_{01},a_1)$ is 0. That means the robot call a passerby that is passing by it will get no value.}
    \label{fig:exqfunc}
\end{figure}

We visualize the Q-function before the learning and the Q-function after the learning by heat map in order to analyze the effect of UCQL.
UCQL can change the action by updating Q-function.
Therefore, we can know how robot learn the action by visualizing Q-function.
Figure \ref{fig:exqfunc} is an example Q-function to explain a visualization on this paper.

\section{Result\label{cap:result}}
We constructed a experiment environment described on Method in the entrance of our buildings.
Figure \ref{fig:example} shows a picture of the equipment in the environment.
The experimenter was the corresponding author.
The participants were a lot of employees and visitors of our company. The learning interval is three days. 
As a result, we measured a lot of data. We clean the data by the following step because the data have a lot of noise on the field such as detection errors by Motion Capture and so on.
\begin{itemize}
\item We drop episodes that interval is less than 1 [sec] because it takes a 3 [sec] to walk across the detection area of Motion Capture.
\item We drop episodes that is from $s_{00}$ to $s_{00}$ only because nobody was in the detection area of Motion Capture.
\end{itemize}

We got 209 total episodes in the experiment after the data cleansing.
Table \ref{tab:params} shows number of episodes and time on each condition.
We calculated the accuracy from the confusion matrix on each condition.
The confusion matrices for the before condition and the after condition were respectively $(\textrm{TP,FP,FN,TN})=(11,59,0,17)$ and $(\textrm{TP,FP,FN,TN})=(7,23,0,92)$.
Therefore, the accuracy of the baseline and proposed methods were respectively 0.322 and 0.811. In testing $WH$ by the one-sided Test of Proportion, we found a significant difference in accuracy between the before and after condition ($p=4.46 \times 10^{-13} < 0.01$).

%I observed a passerby approaches the robot when robot do $a_6$.

% After Precision:0.25 TP:1 FP:3 4.5h 1626
%2nd day Precision:0.2222222222222222 TP:2 FP:7 5.1h 1800
% Before 0.11764705882352941 TP:4 FP:30 4.2h 2605
%TP:4 FP:31 4.1h 2572

%1000msec over
%After 1st:  TP:1 FP:2 14
%2nd: TP:2 FP:6 25
%3rd: TP:0 FP:1 3
%4th: TP:1 FP:6 37
%4th: TP:1 FP:2 12
%Before 1st: TP:4 FP:24 38
%2nd: TP:4 FP:28 37

%http://www.tablesgenerator.com/latex_tables

\begin{table}[tb]
    \caption{ Items of the result after the data cleansing.}
    \label{tab:params}
    \centering
    \begin{tabular}{|l||r|r||r|}
        \hline
        items          & Before& After & Total \\ \hline
        episodes       &   87   & 122      & 209   \\ \hline
        time{[}h{]}   &    13.7   & 26.7       & 40.4    \\ \hline
        days{[}d{]}    &    3   & 6       & 9    \\ \hline
    \end{tabular}
\end{table}

\begin{figure}[tb]
    \centering
    \includegraphics[width=0.5\linewidth]{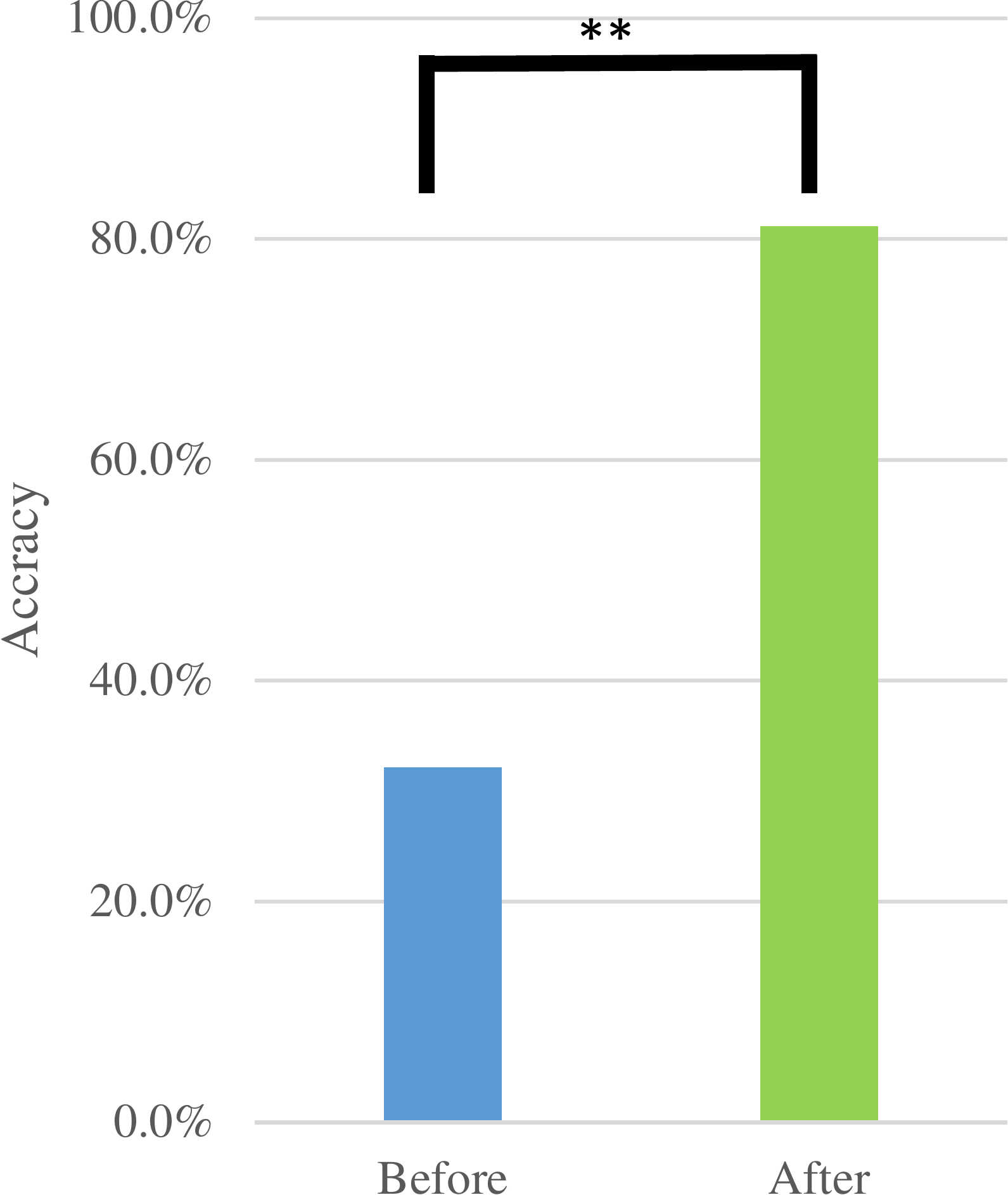}
    \caption{The accuracy of the experiment on each condition (**: $p<0.01$)}
    \label{fig:expaccuracy}
\end{figure}

\begin{figure*}[t]
    \begin{minipage}{0.48\hsize}
        \centering
        \includegraphics[width=0.98\linewidth]{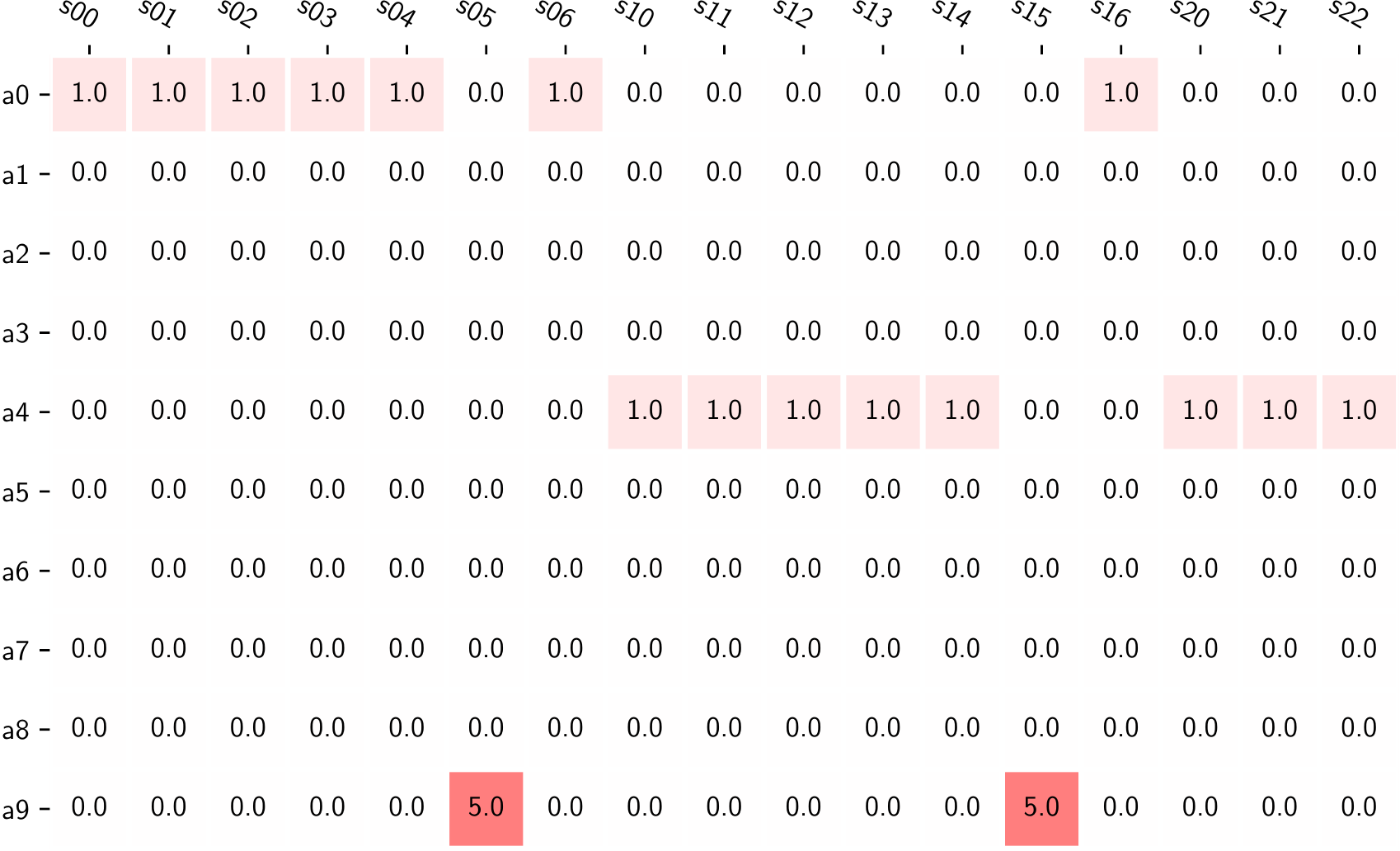}
        \subcaption{The part of the Q function before the learning ($Q_B$).}
        \label{fig:exppqresult_before}
    \end{minipage}
    \begin{minipage}{0.48\hsize}
        \centering
        \includegraphics[width=0.98\linewidth]{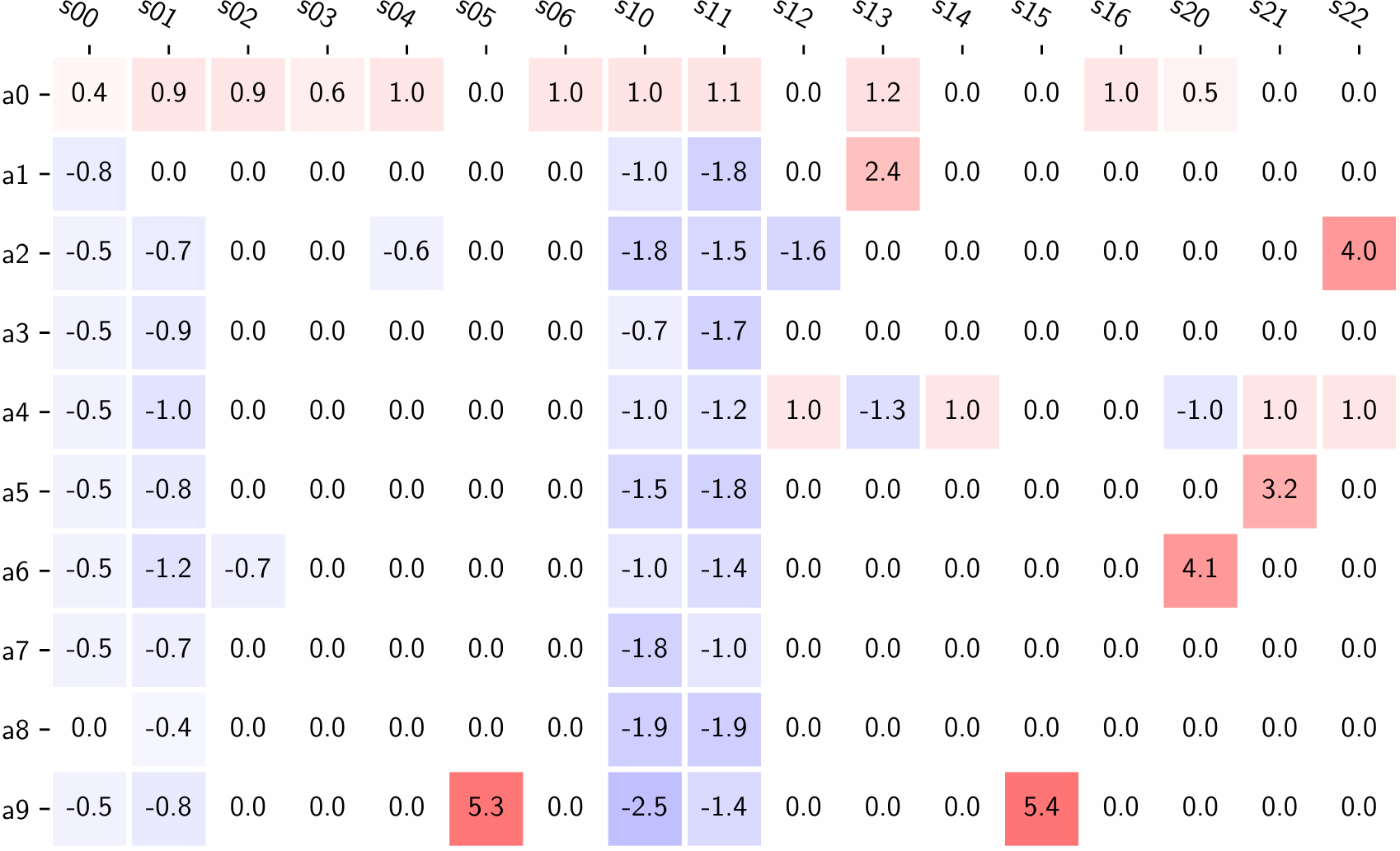}
        \subcaption{The part of the Q function after the learning ($Q_A$).}
        \label{fig:exppqresult_after}
    \end{minipage}
    \caption{The changing process of Q-function by UCQL.}
    \label{fig:exppqresult}
\end{figure*}

\section{Discussion}
\label {Discussion}
We discuss the original hypothesis, "The robot can attract passersby without users' discomfort by User-Centered Reinforcement Learning.", in the point of following views.

\begin{enumerate}
\item Can we accept the original hypothesis?
\item Why the robot attract passersby without discomfort by the proposed method?
\item What is the limitations of the method and the experiment?
\end{enumerate}

\subsection{Can we accept the original hypothesis?}
We explain why we can accept the original hypothesis by using the result of the experiment and another study.

At first, we show that the we can accept $WH$, "The accuracy after a learning by UCQL is better than the accuracy before a learning by UCQL".
According to Capture \ref{cap:result}, we found a significant difference in precision between the before and after condition. Thus, we accept $WH$. Therefore, we can infer $WH$ as true.

The result of the experiment supports the original hypothesis though the above-mentioned discussion because the working hypothesis is true.
Therefore, we can accept the original hypothesis.

\subsection{Why the robot attract passersby without discomfort by the proposed method?}
We can explain why the robot attract passersby without discomfort in view of the learning process with Figure. \ref{fig:exppqresult}.

Why the robot reduce FN by UCQL? We compare the row of $s_{01}$ in Figure. \ref{fig:exppqresult}(a) and the row of $s_{01}$ in Figure. \ref{fig:exppqresult}(b). The robot before learning selected a action $a_4$ because $\argmax_a Q_B(s_{01},a)=a_4$. The robot after learning selected a action $a_0$ because  $\argmax_a Q_A(s_{01},a)=a_0$. That means robot do not calls if passerby don't use a robot service. Therefore, the robot reduce FN by UCQL.

\subsection{What is the limitations of the method and the experiment?}
In this experiment, we supposed that a passerby do not walk with others. In other words, we do not consider a group of passersby. Thus, we need to expand the method in order to process a group of them.

The data in this study are sampled from biased population. We need to take further experiments on other environments if we want more soundness about the working hypotheses.

In this experiment, we create the reward function based on other studies. However, it is hard to create reward functions on each case. Therefore, we have to create a easy method in order to design reward function and Q function.

\section{Conclusion}
\label {conclusion}
We investigated the hypothesis that "by using user-centered Q-learning, a robot can increase the chance of being able to provide a service to a passerby without causing the passerby discomfort." We proposed a method based on reinforcement learning in robotics and focused on the reward function and the Q-function because we wanted the robot to perform actions in view of user experience?. To investigate our hypothesis, we made a working hypothesis and tested it experimentally. From the results, we accepted the working hypothesis and the original hypothesis.

Future work includes generalizing the method for creating the reward function to make it applicable to different tasks and developing a distributed reinforcement learning method that enhances time-efficiency.

\bibliographystyle{IEEEtran}
\bibliography{ms}

\end{document}